%% file: master.tex
\pdfoutput=1
\documentclass[]{article}
\usepackage{ijcai21}

\usepackage{times}

\usepackage{soul}
\usepackage{url}
\usepackage[hidelinks]{hyperref}
\usepackage[utf8]{inputenc}
\usepackage[small]{caption}
\usepackage{graphicx}
\usepackage{amsmath}
\usepackage{booktabs}
\usepackage{float}
\usepackage{caption}
\input{my_macros}

\title{\mytitle}

\author{
Shusaku Sone$^1$\and
Jiaxin Ma$^1$\and
Atsushi Hashimoto$^1$\footnote{Contact Author,Corresponding Author}\and
Naoya Chiba$^{1,2}$\And
Yoshitaka Ushiku$^1$
\\
\affiliations
$^1$OMRON SINIC X Corporation, $^2$Tohoku University\\
\emails
\{atsushi.hashimoto,shusaku.sone\}@sinicx.com}

\begin{document}

\maketitle

\begin{abstract}
\input{sections/0_abst}
\end{abstract}

\input{sections/1_introduction}

\input{sections/2_relatedwork}

\input{sections/3_network_architecture}
\input{sections/4_experiments}

\input{sections/6_conclusion}

\newpage
\bibliographystyle{named}
\bibliography{my_bib}

\input{sections/Z_supplementary}

\end{document}

%% file: my_macros.tex
\usepackage{comment}
\usepackage{subfigure}
\usepackage{amsmath}
\usepackage{bm} 
\usepackage{ascmac} 
\usepackage{multirow} 

\usepackage{booktabs}
\usepackage{bbding}
\usepackage{pifont}
\usepackage{wasysym}
\usepackage{amssymb}

\newcommand{\mytitle}{WeaveNet for Approximating Two-sided Matching Problems}

\makeatletter
\def\WordCount#1{%
\@tempcnta\z@%
\@tfor \@tempa:=#1\do{\advance\@tempcnta\@ne}%
\@tempcnta%
}%
\makeatother

\newif\ifprintjp

\newcommand{\para}[2]{%

\ifprintjp%
  \ifx"#1"
  #2
  \else%
  #1 
  \fi
\else%
  \ifx"#2"
  #1
  \else%
  #2 
  \fi
\fi

}

\usepackage[dvipsnames]{xcolor}

\newcommand{\writtenby}[1]{}
\newcommand{\checkedby}[2]{}

\usepackage{url}

\newcommand{\loss}{{\mathcal L}}
\newcommand{\lowest}{C_{min}}

\newcommand{\SEq}{S\hspace{-0.1em}Eq}

%% file: sections/0_abst.tex
Matching, a task to optimally assign limited resources under constraints, is a fundamental technology for society.
The task potentially has various objectives, conditions, and constraints; however, the efficient neural network architecture for matching is underexplored.
This paper proposes a novel graph neural network (GNN), \textit{WeaveNet}, designed for bipartite graphs.
Since a bipartite graph is generally dense, general GNN architectures lose node-wise information by over-smoothing when deeply stacked.
Such a phenomenon is undesirable for solving matching problems.
WeaveNet avoids it by preserving edge-wise information while passing messages densely to reach a better solution.
To evaluate the model, we approximated one of the \textit{strongly NP-hard} problems, \textit{fair stable matching}.
Despite its inherent difficulties and the network's general purpose design, our model reached a comparative performance with state-of-the-art algorithms specially designed for stable matching for small numbers of agents.

%


%% file: sections/1_introduction.tex
\section{Introduction}
From job matching to multiple object tracking (MOT),  matching problems can represent various decision-making applications.
A matching problem is generally described on a bipartite graph, a graph that has two sets of nodes $A$ and $B$ with edges $E=A\times B$ ($N=|A|$, $M=|B|$, $N\geq M$).
On the graph, the task is to find a matching $m$ (a set of edges that do not share any nodes) that maximally satisfies an objective.
The most famous objective is to maximize the sum of edge weights in $m$, which can be solved by the Hungarian algorithm \cite{kuhn1955hungarian}. This problem is called {\it one-sided} because only one stakeholder dominates both sides $A$ and $B$. In contrast, {\it two-sided matching} supposes that agents on each side aim to maximize their reward.

%

A fairer two-sided matching solver provides more agreeable social decisions, but some fairness objectives make the problem {\it strongly NP-hard} \cite{kato1993complexity,mcdermid2014sex}.
Besides, agents may randomly appear/disappear in dynamic matching problems (e.g., MOT with occlusions \cite{MOT_match}).
In such cases, we need to compensate for the inputs of incomplete information by its stochastic properties.
A learning-based model can be an alternative to hand-crafted algorithms for such applications.
However, such models are underexplored.

This paper aims to propose an effective learning-based solver for matching problems and evaluate it on {\it strongly NP-hard} conditions of {\it fair stable matching problems}.

The contribution of this paper is four-fold:
\begin{enumerate}
   \item We proposed {\it WeaveNet}, a novel neural network architecture for matching problems that largely outperforms traditional learning-based approaches.
   \item For the first time, our learning-based solver achieved a comparative performance to the state-of-the-art handcrafted algorithm of NP-hard matching problems, though it was achieved with the limited size of problem instances ($N=20,~30$).
   \item For the first time, we demonstrated the performance of a deep model unsupervisedly trained by a continuous relaxation of discrete conditions of matching problems.
   \item We provided a steady baseline for the problem with source code to open a new vista for learning-based combinatorial optimization\footnote{The source code and datasets are included in this submission and will be publicly available.}.
\end{enumerate}

%% file: sections/2_relatedwork.tex
\section{Background and Related Work}

\subsection{Stable Matching Problem}\label{ss:smp}
\begin{figure}[ht!]
    \centering
    \includegraphics[width=1.0\linewidth]{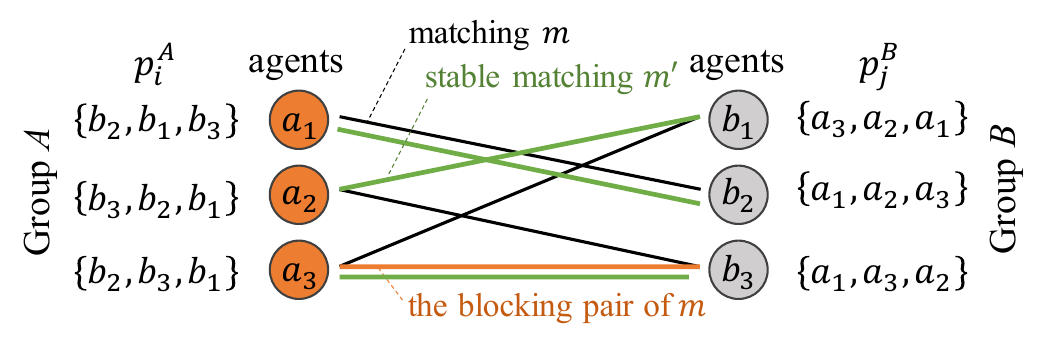}
    \caption{
    An example of a stable matching instance with a size of $N=3$. The matching $m$ (black edges) is not stable due to the blocking pair (the orange edge). The matching $m'$ (green edges) is a stable matching.}
    \label{fig:stable_matching}
\end{figure}

In this study, we evaluate our model with a typical two-sided problem of stable matching.
An instance $I$ of a stable matching problem consists of two sets of agents $A$ and $B$ on a bipartite graph. 
Fig.~\ref{fig:stable_matching} illustrates an example of $I$. 
Each agent $a_i$ in $A~(0<i\leq N)$ has a preference list $p^A_i$, which is an ordered set of elements in $B$ and $p^A_{ij}=rank(b_j;p^A_i)$ is the index of $b_j$ in the list $p^A_i$.
$a_i$ prefers $b_j$ to $b_{j'}$ if $p^A_{ij}<p^A_{ij'}$.
Similarly, each agent $b_j$ in $B (0<j\leq M)$ has a preference list $p_j^B$.

For a matching $m \in \{0,1\}^{N\times M}$, we say that an unmatched pair $\{a_v, b_w\}(\notin m)$ blocks $m$ if there are matched pairs $\{a_v,b_j\}\in m$ and $\{a_i,b_w\}\in m$ that satisfy $p^A_{vw} < p^A_{vj}$ and $p^B_{wv} < p^B_{wi}$. Here, $\{a_v, b_w\}$ is called a blocking pair (the orange edge blocks a matching of black edges in the figure).

A matching is called stable if it has no blocking pair (the green edges in the figure).
It is known that an instance $I$ always has at least one stable matching, and the Gale-Shapley (GS) algorithm can find it in $O(N^2)$ \cite{gale1962college}.
However, the algorithm prioritizes either side and the other side only gets the least preferable result among all the possibilities of stable matching.

\begin{table}[ht]
   \small
   \setlength\tabcolsep{4.0pt} 
    \centering
    \begin{tabular}{lrcl}
        \toprule
        \multicolumn{1}{c}{Cost} & \multicolumn{3}{c}{Definition} \\
        \midrule
        Sex equality  &  $\SEq(m;I)$ & $\!=\!$  & $|P(m;A)-P(m;B)|$ \\
        Regret        &  $Reg(m;I)$ & $\!=\!$  & $\max_{\{a_i,b_j\}\in m}(\max(p^A_{ij},p^B_{ji}))$ \\
        Egalitarian   &  $Egal(m;I)$& $\!=\!$  & $P(m;A)+P(m;B) $ \\
        Balance       &  $Bal(m;I)$ & $\!=\!$  & $\max(P(m;A),P(m;B)) $  \\   \bottomrule
    \end{tabular}
    \caption{Representative costs to measure the fairness of matching.}
    \label{tab:costs}
\end{table}


To compensate for the unfairness, there have been diverse measures of fairness proposed, with different complexities of finding an optimal solution. They are listed in Table \ref{tab:costs} with the definitions, where
\begin{align}
    P(m;A) =\!\! \sum_{\{a_i,b_j\}\in m}\!\! p^A_{ij}, && P(m;B) =\!\! \sum_{\{a_i,b_j\}\in m}\!\! p^B_{ji}.  \label{eq:pref} 
\end{align}
\textbf{Sex equality cost} ($\SEq$) is a popular measure that focuses on the unfairness brought by the gap between the two sides' satisfaction \cite{gusfield1989stable}. 
\textbf{Regret cost} ($Reg$) measures the unfairness brought by the agent who obtained the least satisfaction (the weakest individual) in the matching.
\textbf{Egalitarian cost} ($Egal$) measures the overall satisfaction, in other words, the greater good, rather than concerning the weaker individual or weaker side \cite{gusfield1989stable}.
\textbf{Balance cost} ($Bal$) is a compromise between side-equality and overall satisfaction \cite{feder1995stable,gupta2019balanced}, and equivalent to $\frac{1}{2}(\SEq+Egal)$. 
Minimizing balance cost is to benefit the weaker side while less harming the stronger side. 

Among these measures, a stable matching of minimum regret cost and minimum egalitarian cost can be found in $O(N^2)$ and $O(N^3)$, respectively \cite{gusfield1987three,irving1987efficient,feder1992new}, while minimizing sex equality cost or balance cost is known as strongly NP-hard \cite{kato1993complexity,mcdermid2014sex,feder1995stable}.

There are several approximation solvers for the strongly NP-hard stable matching variants. Iwama \emph{et~al$.$} proposed a method that runs in $O(N^{3+\frac{1}{\varepsilon}})$ whose gap from the optimal solution is bounded by $\varepsilon$ \cite{iwama}.
Deferred Acceptance with Compensation Chains 
(\textit{DACC}) is an extension of the GS algorithm without prioritizing any side. 
It can terminate in $O(N^4)$ \cite{dworczak2016deferred}. 
\textit{PowerBalance} is a state-of-the-art algorithm.
In each round of proposing, it balances the two sides by letting the stronger side propose (because their satisfactions decrease if getting rejected). 
It was reported that PowerBalance has competitive performance when enforcing a termination in $O(N^2)$ \cite{tziavelis2019equitable}.

\subsection{Learning-based Matching Solvers}\label{ss:mlap}
Although learning-based matching solvers are still in the process of development, it has potential applications when we can only observe incomplete information.
The dynamic matching problem \cite{wang2019adaptive} is one typical case.
Another case is temporal alignment, where we have only superficial observation (videos, audio, or narration) and cannot access the true similarity between temporal events. Learning-based solvers would be helpful to jointly train the feature encoder with the matching module \cite{fathony2018efficient}.

Learning-based solvers also have advantages in flexibility even to deal with variations of traditional matching problems (e.g., incomplete preference lists with ties, continuous preference scores, or multi-dimensional preference \cite{multi_stable_matching}) as well as more complex objective functions (e.g., a weighted sum of costs in Table \ref{tab:costs}).
Aiming to develop learning-based methods for the stable matching problem, \cite{harvard2019} proposed a loss function that relaxes the discrete constraints on matching stability into a continuous measure, which enables to train a model in an unsupervised manner.
However, they only provide an experiment on $N=5$ with a shallow neural network.

Deep Bipartite Matching (DBM) \cite{gibbons2019deep} is another attempt to approximate the (not strongly) NP-hard problem of weapon-target assignment (WTA), but its performance is not at a comparative level with the state-of-the-art of WTA approximation
\cite{ahuja2007exact} even limiting the size of problem instances.

%% file: sections/3_network_architecture.tex

\begin{figure*}[!th]
    \centering
    \includegraphics[width=1.0\linewidth]{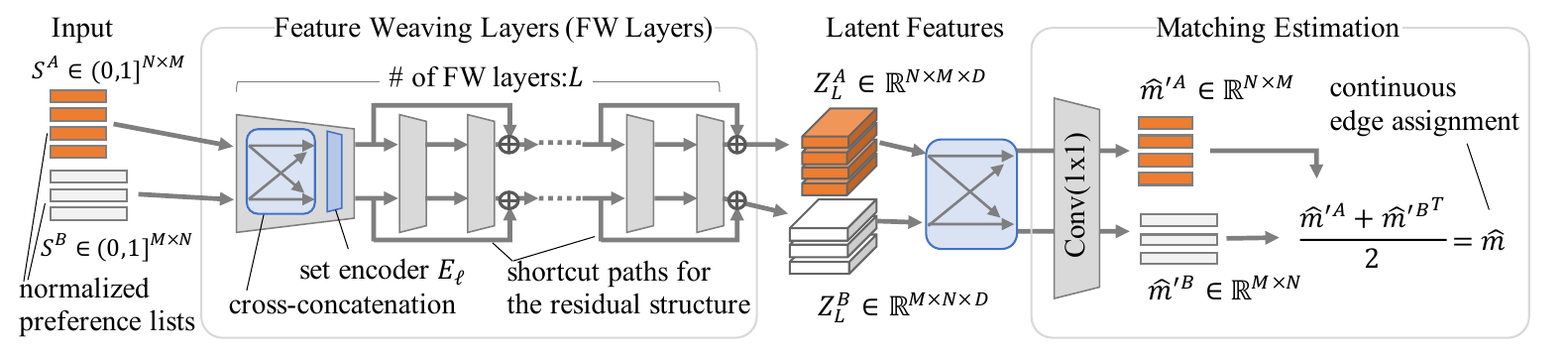}
    \caption{WeaveNet architecture. $L$ feature weaving layers are stacked with shortcut paths to be a deep network. The encoded features are fed into Conv(1$\times$1) layer to obtain logits ($\hat{m}'^A$, $\hat{m}'^B$). The output $\hat{m}$ will be binarized in prediction phase to represent a matching.}
    \label{fig:weavenet}
\end{figure*}

\section{Deep-learning-based Fair Stable Matching}
\writtenby{Hashi}

%
\subsection{WeaveNet}\label{ss:weavenet}
\subsubsection{Input and Output Data Format}
We aim to realize a trainable function $F$ that outputs a matching $\hat{m}$, which is an $N\times M$ matrix.
As for the input, we firstly linearly re-scale\footnote{The details of this linear re-scaling are based on \cite{harvard2019} and described in \ref{ss:sf}. Note that sections numbered with capital letters appear in the supplementary material.} the rank of preference $p^*_{ij}~(*\in \{A, B\})$ ranged in $[1, N]$ into a normalized score $s^*_{ij}$ ranged in $(0, 1]$ to make it invariant to $N$, where 1 for the highest rank.
Then, we obtain the input as matrices $S^A$ and $S^B$, where $s^A_{ij}$ is the $ij$-element of $S^A$.

\subsubsection{Requirement, Intuition and Entire Model}
One of the required properties of $F:(S^A,~S^B)\rightarrow \hat{m}$ is to take all the agents' preference into account when determining the presence of each edge in the output $\hat{m}$. 
\cite{harvard2019} implemented this by multilayer perceptron (MLP), 
where
$S^A$ and $S^B$ are destructured and concatenated into a flat vector (with the length of $2N\hspace{-0.1em}M$) and fed to the MLP. Its output (a flat vector with the length of $N\hspace{-0.1em}M$) is restructured into a matrix $\hat{m}\in {\mathcal R}^{N\times M}$.
The MLP model, however, would face difficulties due to the following four problems.
\setlength{\leftmargini}{15pt}{
\begin{description}
	\setlength{\itemsep}{0pt}      
	\setlength{\parskip}{0pt}      
	\setlength{\itemindent}{-15pt}   
	\setlength{\labelsep}{3pt}     
 \item[(a)] Preference lists of multiple agents are encoded by independent parameters, though they share a format so that we could efficiently process them in the same manner.
 \item[(b)] MLP only supports a fixed-size input, so training different models for different cases of $N$ becomes mandatory.
 \item[(c)] $F$ should be permutation invariant, which means the matching result unchanged even if we shuffle the order of agents in $S^A$ and $S^B$, but MLP does not satisfy.
 \item[(d)] A shallow MLP model may be insufficient to approximate an exact solver for the NP-hard problem when $N$ is large.
\end{description}
}

To address the above weaknesses of MLP, we propose the feature weaving network ({\bf WeaveNet}) which has the properties of (a) {\bf shared encoder}, (b) {\bf variable-size input}, (c) {\bf permutation invariance}, and (d) {\bf residual structure}. The WeaveNet, as shown in Fig.~\ref{fig:weavenet}, consists of $L$ feature weaving (FW) layers. It has two streams of $A$ and $B$. In a symmetric manner, each stream models the agent's act of selecting the one on the opposite side while sharing weights to enhance the parameter efficiency. The shortcut paths at every two FW layers make them residual blocks, which allows the model to be as deep as possible. We explain its details as follows.

\begin{figure}[!ht]
    \centering
    \includegraphics[width=1.0\linewidth]{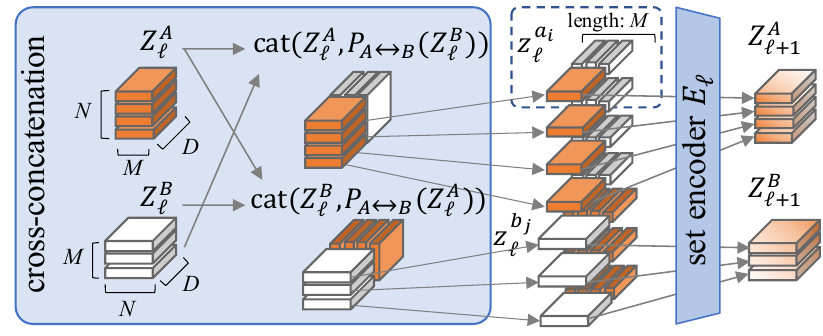}
    \caption{Feature weaving layer orthogonally concatenates the weftwise and warpwize components ($Z_\ell^A$ and $Z_\ell^B$) in a symmetric way (cross-concatenation). Then, the concatenated tensors are separated into $z_\ell^{\smash{a_i}}$ (or $z_\ell^{\smash{b_j}}$), which represents a set of outgoing edges from agent $a_i$ (or $b_j$), and independently fed to $E_\ell$.}
    \label{fig:feature_weaving_layer}
\end{figure}

\subsubsection{Feature Weaving Layer}
Fig.~\ref{fig:feature_weaving_layer} illustrates the detail of a single FW layer, which is the core architecture of the proposed network.
FW layer is a two-stream layer whose inputs consist of a {\it weftwise} component $Z^A_{\ell}$ and a {\it warpwise} component $Z^B_{\ell}$, where $Z^A_0=S^A$ and $Z^B_0=S^B$. 
The two components are symmetrically concatenated in each stream ({\bf cross-concatenation}). Then these concatenations are separated into agent-wise features, each of which is a set of outgoing-edge features of an agent (indicating the preference from that agent to every matching candidate).
These features are processed by the encoder $E_\ell$ {\bf shared by every agent in both $A$ and $B$}.
As for an encoder that can embed {\bf variable-size} input in a {\bf permutation invariant} manner,
we adopted the structure proposed in DeepSet \cite{deepset} and PointNet \cite{pointnet}, which consists of two convolutional layers with kernel size $1$ and a set-wise max-pooling layer, followed by batch-normalization and activation.
We refer to this structure as {\it set encoder}.
See \ref{ss:set_encoder} for visualization of the set encoder and its calculation cost.

\subsubsection{Difference from DBM}
Note that DBM \cite{gibbons2019deep} also satisfies the requirements (b), (c), and (d). The critical difference between WeaveNet and DBM is its parameter efficiency and the choice of the local structure $E_\ell$. DBM applies the weftwise and warpwise communications alternately in a single-stream architecture. Hence, each encoder is specialized in each direction. In our model, every encoder is trained for both directions, and thus it is twice parameter efficient.
See \ref{app:scope} to know how WeaveNet and DBM propagate the weftwise and warpwise communications.
Also, the local structure of DBM is sub-optimal to encode the relative identity of each outgoing edge among $N\times M$ pairs. 
We show the difference in performance brought by the local structure $E_\ell$ in the experiment section.

\checkedby{Jiaxin}{Hashi}
\subsubsection{Mathematical Formulations of WeaveNet}
The input $Z^A_\ell$ is a third-order tensor whose dimensions, in sequence, corresponding to the agent, candidate, and feature dimension, with a size of $(N, M, D)$.
Similarly, $Z^B_\ell$ has a size of $(M, N, D)$.
The {\bf cross-concatenation} is defined by function $P_{A\leftrightarrow B}$, which swaps the first and second dimensions of the tensor, and $cat(\{Z_1,Z_2,\ldots\})$, which concatenates the features of two tensors $Z_1,~Z_2$, as
\begin{equation}
\begin{split}
 Z'^A_\ell &= cat(Z^A_\ell,P_{A\leftrightarrow B}(Z^B_\ell)).
\end{split}
\end{equation}
$Z'^A_{\ell}$ is sliced into agent-wise features $z^{a_i}_\ell$ and we obtain $Z^A_{\ell+1} = (E_\ell(z^{a_i}_\ell)|0<i\leq N)$. We can calculate $Z^B_{\ell+1}$ in a symmetric manner (with the same encoder $E_\ell$).

\checkedby{Jiaxin}{Hashi}
After the process of $L$ FW layers, $Z^A_L$ and $Z^B_L$ are further cross-concatenated and fed to the matching estimator (in Fig.~\ref{fig:weavenet}). It outputs a non-deterministic matching $\hat{m}$.
In the training phase, $\hat{m}$ is input to an objective function, and the loss is minimized.
In the prediction phase, the matching is obtained by binarizing $\hat{m}$.
In this sense, matching estimation through a neural network can be considered as an approximation by relaxing the binary matching space $\{0,1\}^{N\times M}$ (where 1 represents the edge selection while 0 for rejection) into a continuous matching space $[0,1]^{N\times M}$.

\subsubsection{Symmetric Property and Asymmetric Variant}
WeaveNet is designed to be fully symmetric for $S^A$ and $S^B$. Hence, it satisfies the equation $F(S^A, S^B)=F(S^B, S^A)^\top$.
This condition ensures that the model architecture cannot distinguish the two sides $A$ and $B$ innately. 
This property is beneficial when mathematically fair treatment between $A$ and $B$ is desirable. 
However, when inputs from $A$ and $B$ are differently biased (e.g., the two sides have different trends of preference), sharing encoders in two directions may degrade the performance.
To eliminate the bias difference without losing the parameter-efficiency, we further propose to \textbf{a)} apply batch normalization independently for each stream, and \textbf{b)} adding a side-identifiable code (e.g., 1 for $A$ and 0 for $B$) to $Z^A_0$ and $Z^B_0$ as a ($D$+1)-th element of the feature. We call this variant ``asymmetric''.

\subsection{Relaxed Continuous Optimization}\label{ss:relaxation}
Generally, a combinatorial optimization problem has discrete objective functions and conditions, which are not differentiable. To optimize the model in an end-to-end manner without inaccessible ground truth, we optimize the model by relaxing such discrete loss functions into continuous ones. 

Assume we target to obtain a fair stable matching that has the minimum $\SEq$, for example. Then, we have the following three loss functions.
\setlength{\leftmargini}{15pt}{
\begin{description}
	\setlength{\itemsep}{0pt}      
	\setlength{\parskip}{0pt}      
	\setlength{\itemindent}{-15pt}   
	\setlength{\labelsep}{3pt}     
 \item[$\loss_m$] conditions the binarization of $\hat{m}$ to represent a matching.
 \item[$\loss_s$] conditions the matching to be stable.
 \item[$\loss_f$] minimizing the fairness cost $\SEq$ of the matching
\end{description}
}
The overall loss function is defined as
\begin{align}
\hspace{-0.5em}    \loss_{\rm fsm}(\hat{m})&= \lambda_m\loss_m + \frac{1}{2}\hspace{-1.0em}\sum_{m\in \{\hat{m}^A,\hat{m}^B\}}\hspace{-1.3em}\bigl( \lambda_s\loss_s(m) + \lambda_f\loss_{f}(m)\bigr), \label{eq:lfsm}
\end{align}
where $\hat{m}^A = {\rm softmax}(\hat{m})$ and $\hat{m}^B = {\rm softmax}(\hat{m}^{\top})$. 

An advantage of learning-based approximation is its flexibility. We can modify the above loss functions to easily obtain other variants. For example, removing $\loss_{f}$ in Eq.\eqref{eq:lfsm} leads to standard stable matching, and replacing $\loss_{f}$ with $\loss_{b}$ (which minimizes $Bal$) leads to balanced fair-stable matching, as follows:
\begin{align}
\hspace{-0.5em}    \loss_{\rm sm}(\hat{m}) &= \lambda_m\loss_m \label{eq:lsm} + \frac{1}{2}\hspace{-1.0em}\sum_{m\in \{\hat{m}^A,\hat{m}^B\}}\hspace{-1.3em}\lambda_s\loss_s(m),\\
\hspace{-0.5em}    \loss_{\rm bsm}(\hat{m}) &=  \lambda_m\loss_m + \frac{1}{2}\hspace{-1.0em}\sum_{m\in \{\hat{m}^A,\hat{m}^B\}}\hspace{-1.3em}\bigl(\lambda_s\loss_s(m) + \lambda_b\loss_{b}(m)\bigr). \label{eq:lbsm}
\end{align}

\subsubsection{Matrix Constraint Loss \texorpdfstring{$\loss_m$}{Lm}}
$\hat{m}$ can be safely converted into a binarized matching by column-wise or row-wise ${\rm argmax}$ operation when it is a symmetric doubly stochastic matrix \cite{harvard2019}.
To satisfy this condition, we defined $\loss_m$ with an average of the cosine distance as
\begin{equation}
\begin{split}
\hspace{-0.5em} {\rm C}(\hat{m}^A,\hat{m}^B)&=\frac{1}{N}\sum_{i=0}^N\frac{\hat{m}^A_{i*}\cdot\hat{m}^B_{*i}}{\|\hat{m}^A_{i*}\|_2\|\hat{m}^B_{*i}\|_2}\\
\hspace{-0.5em} \loss_m(\hat{m}^A,\hat{m}^B)&=1-\frac{({\rm C}(\hat{m}^A,\hat{m}^B)+{\rm C}(\hat{m}^B,\hat{m}^A))}{2},
\end{split}
\label{eq:loss_c_p2}
\end{equation}
where $\hat{m}^A_{i*}$ means the $i$-th row of $\hat{m}^A$.
This formulation binds $\hat{m}$ to be a symmetric\footnote{Here the symmetry of $\hat{m}$ is defined by $\hat{m}_{i*} = \hat{m}_{*i}, i=1...M$ in case that $\hat{m}$ is not a square matrix ($N>M$).} doubly stochastic matrix when $\loss_m(\hat{m}^A,\hat{m}^B)=0$.
The advantage of this implementation against the original one in \cite{harvard2019} is described in \ref{ss:exp1_supp}.

\subsubsection{Blocking Pair Suppression by \texorpdfstring{$\loss_s$}{Ls}}
As for $L_s$, we used the function proposed in \cite{harvard2019} as it is, which is 
\begin{equation}
\begin{split}
g(a_i;b_w,\hat{m}) &= \sum_{b_j\neq b_w}\hat{m}_{ij}\cdot \max(S^A_{iw}-S^A_{ij},0) \\ 
g(b_j;a_v,\hat{m}) &= \sum_{a_i\neq a_v}\hat{m}^{\top}_{ji}\cdot \max(S^B_{jv}-S^B_{ji},0) \\ 
\loss_s(\hat{m};I) &= \hspace{-1.0em}\sum_{(v,w)\in A \times B}\hspace{-1.0em}g(a_v;b_w,\hat{m})g(b_w;a_v,\hat{m}), 
\end{split}
\end{equation}
where $g(a_i;b_w,\hat{m})$ is a criterion known as ex-ante justified envy, which has a positive value when $a_i$ prefers $b_w$ more than any $b_j$ in $\{b_j|j\neq w,\hat{m}_{ij}>0\}$. This is the same for $g(b_j;a_v,\hat{m})$.
Hence, $\{a_v,b_w\}$ becomes a (soft) blocking pair when both $g(a_v;b_w,\hat{m})$ and $g(b_w;a_v,\hat{m})$ are positive. 

\subsubsection{\texorpdfstring{$\loss_{f}$ and $\loss_{b}$}{Lf and Lb} as Fairness Measurements}
$\loss_f$, $\loss_b$ minimize $\SEq(m;I)$, $Bal(m;I)$ in Table \ref{tab:costs}, respectively, and are defined as
\begin{align}
\loss_f(\hat{m};I) &= \frac{1}{N}|S(\hat{m};A) - S(\hat{m};B) |,\\
\loss_b(\hat{m};I) &= -\frac{1}{N}{\rm min} (S(\hat{m};A),S(\hat{m};B)),
\end{align}
where
\begin{equation}
~ S(\hat{m};A) = \sum_{i=1}^{N}\sum_{i=j}^{M} \hat{m}_{ij} \cdot {S^{A}_{ij}},~ S(\hat{m};B) = \sum_{j=1}^{M}\sum_{i=1}^{N} \hat{m}_{ij} \cdot {S^{B}_{ji}}.
\end{equation}


%% file: sections/4_experiments.tex
%

\begin{figure*}[!ht]
    \centering
  \begin{minipage}{0.31\textwidth}
    \centering
    \includegraphics[width=1.0\linewidth]{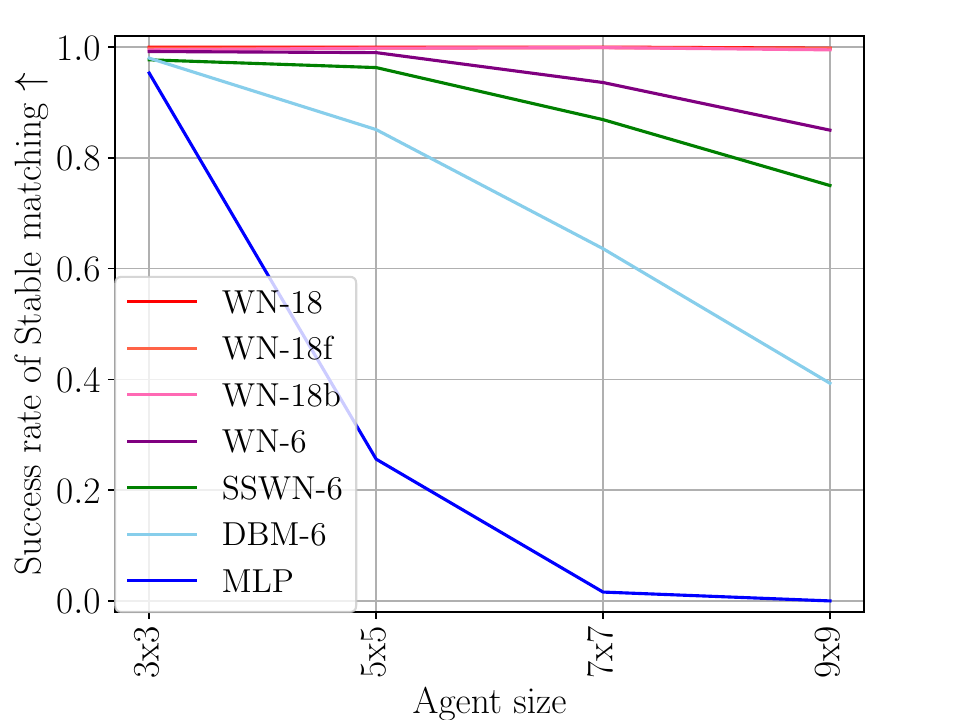}
    \caption{Change in the success rates of stable matching ($\uparrow$) according to $N$.~\newline}
    \label{fig:Ex1_a}
  \end{minipage}%
  \begin{minipage}{0.035\textwidth}~
  \end{minipage}%
  \begin{minipage}{0.31\textwidth}
    \centering
    \includegraphics[width=1.0\linewidth]{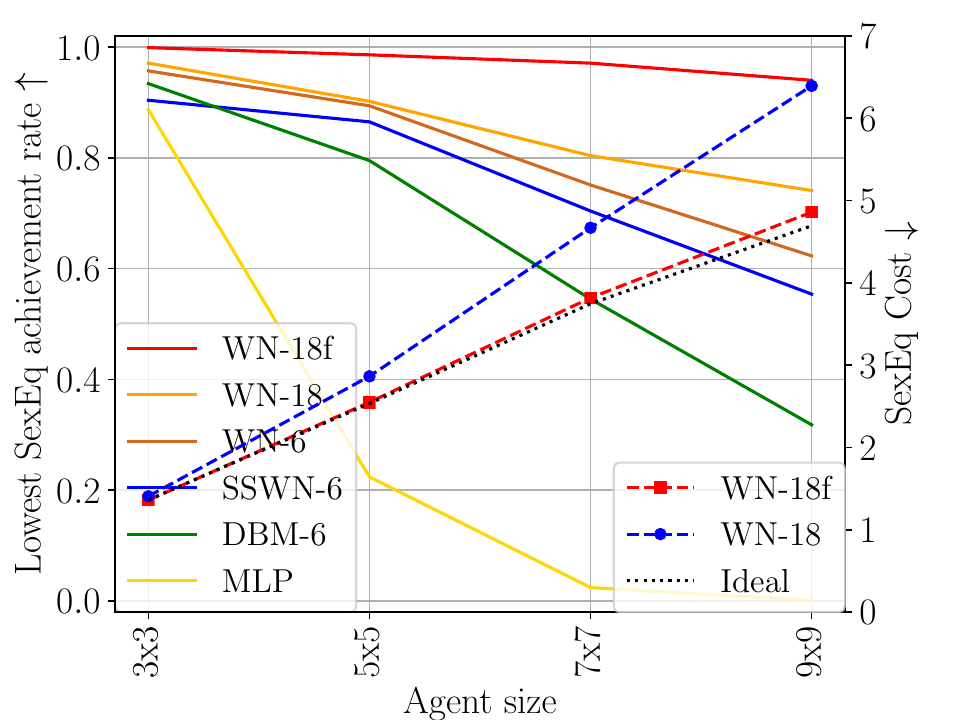}
    \caption{The rates of achieving the ideal fair-stable matching with minimum $\SEq$ (solid, $\uparrow$), and the average $\SEq$ (dashed, $\downarrow$).}
    \label{fig:Ex1_b}
  \end{minipage}%
  \begin{minipage}{0.035\textwidth}~
  \end{minipage}%
  \begin{minipage}{0.31\textwidth}
    \centering
    \includegraphics[width=1.0\linewidth]{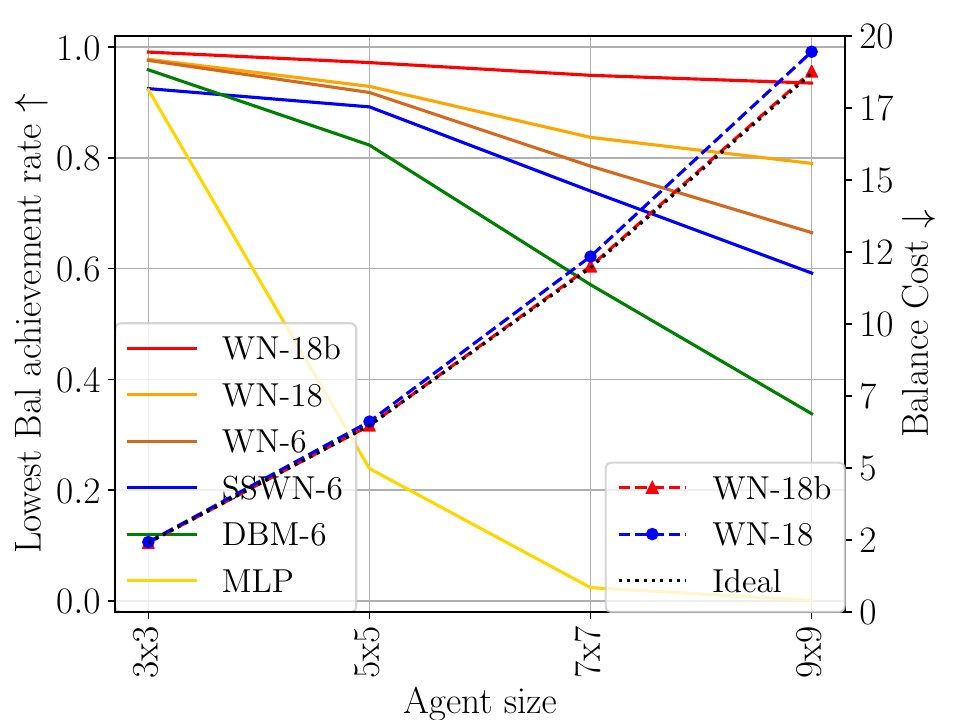}
    \caption{The rates of achieving the ideal fair-stable matching with minimum $Bal$ (solid, $\uparrow$), and the average $Bal$ (dashed, $\downarrow$).}
    \label{fig:Ex1_c}
  \end{minipage}%
\end{figure*}

\section{Experiments}\label{s:exp}
We evaluated WeaveNet through two experiments and one demonstration.
First, with test samples of $N<10$, we compared its performance with learning-based baselines and optimal solutions obtained by a brute-force search.
Second, we compared WeaveNet with algorithmic baselines at $N=20,~30$.
Third, we demonstrate its performance at $N=100$. Note that we always assume $M=N$ hereafter. 

\subsubsection{Sample Generation Protocol}
In the experiments, we used the same method as \cite{tziavelis2019equitable} to generate synthetic datasets that draw preference lists from the following distributions. 
\setlength{\leftmargini}{3pt}{
\begin{description}
\item[Uniform (U)] Each agent's preference towards any matching candidate is totally random, defined by a uniform distribution $\mathcal{U}(0,1)$ (larger value means prior in the preference list).
\item[Discrete (D)] Each agent has a preference of $\mathcal{U}(0.5,1)$ towards a certain group of $\lfloor 0.4N\rfloor$ popular candidates, while $\mathcal{U}(0,0.5)$ towards the rest.
\item[Gauss (G)] Each agent's preference towards $i$-th candidate is defined by a Gaussian distribution $\mathcal{N}(i/N,0.4)$. 
\item[LibimSeTi (Lib)] Simulate real rating activity on the online dating service LibimSeTi \cite{brozovsky07recommender} based on the 2D distribution of frequency of each rating pair ($p^A_{ij},~p^B_{ji}$).
\end{description} 
}

Choosing the above preference distributions for group $A$ and $B$ respectively, we obtained five different dataset settings, namely UU, DD, GG, UD, and Lib.
We randomly generated 1,000 test samples and 1,000 validation samples for each of the five distribution settings.

\subsubsection{Training Protocol}
We trained any learning-based models 200k total iterations at $N\leq 30$ and 300k at $N=100$, with a batch size of $8$\footnote{The number of training samples is $3.7\times 10^{-9}$\% of total cases even when $N=5$ and the overlap with test samples is negligible.}. 
We randomly generated training samples at each iteration based on the distribution of each dataset and used the Adam optimizer \cite{adam}.
We set learning rate 0.0001 and loss weights $\lambda_s=0.7$, $\lambda_m=1.0$, $\lambda_f=\lambda_b=0.01$ based on a preliminary experiment (see \ref{app:hyperparameters}). 

\subsection{Comparison with Learning-based Methods}\label{ss:exp1}

In this experiment, we show results obtained by following baselines and WeaveNet variants.
{\bf MLP} is the model proposed in \cite{harvard2019}. 
{\bf DBM-6} is Deep Bipartite Matching \cite{gibbons2019deep} with $L=6$ layers.
{\bf SSWN-6} is the single-stream WeaveNet with $L=6$ layers.
This variant is to test how WeaveNet works when adopting the single-stream design of DBM.
{\bf WN-6} and {\bf WN-18} is the standard WeaveNet with $L=6$ and $L=18$ layers, respectively.
All the above models were trained to optimize Eq.~\eqref{eq:lsm}. 
{\bf WN-18f} and {\bf WN-18b} have the same architecture with WN-18, but were trained to optimize Eq.~\eqref{eq:lfsm} and Eq.~\eqref{eq:lbsm}, respectively. 
Note that MLP was trained for each $N$ because the network cannot flexibly accept different-size problem instances. The other methods were trained with samples of $N=10$ and tested on samples of $N=3,~5,~7,~9$.
Models with $L=6$ have a similar number of parameters with the MLP model (for sample size $N=5$), so they can have a fair comparison of parameter efficiency. WN models with $L=18$ are aimed to show a larger potential.
See \ref{ss:exp1_supp} for more precise definition of each architecture.

Fig.~\ref{fig:Ex1_a} shows the success rates of finding a stable matching. MLP can hardly find stable matchings when $N\geq 5$. DBM performs better than MLP but still obviously worse than any of WeaveNet variants. Standard WeaveNet's two-stream architecture shows advantages over its single-stream variant.

Fig.~\ref{fig:Ex1_b} and \ref{fig:Ex1_c} show the success rates of finding the possibly fairest (i.e., lowest $\SEq$ or $Bal$) stable matching (solid lines) and the average costs of $\SEq$ and $Bal$ (dashed lines), respectively. Each method has the same trend of performance as in Fig.~\ref{fig:Ex1_a}.
Specifically, the gap between WN-18 and WN-18f/b proved the flexibility of the model for customizing the objective function.

\begin{table*}[ht!]
\setul{1pt}{.4pt}
 \begin{center}
 \setlength\tabcolsep{6pt} 
 \scalebox{1.0}{
 \small
  \begin{tabular}{lrrrrr|rrrrr} \toprule
      Agents ($N\times M$) & \multicolumn{5}{c}{$20 \times 20$ } & \multicolumn{5}{|c}{$30 \times 30$ } \\ 
      Datasets (Distribution Type) & \multicolumn{1}{c}{UU}    & \multicolumn{1}{c}{DD}    & \multicolumn{1}{c}{GG}    & \multicolumn{1}{c}{UD}     & \multicolumn{1}{c}{Lib}    & \multicolumn{1}{|c}{UU}    & \multicolumn{1}{c}{DD}    & \multicolumn{1}{c}{GG}    & \multicolumn{1}{c}{UD}     & \multicolumn{1}{c}{Lib} \\ \midrule
      GS                 & 41.89 & 18.81 & 19.52 & \ul{\bf 70.97}  & 19.66 & 94.03 & 43.46 & 36.56 & \ul{163.77} & 39.78  \\
      PolyMin            & 19.93 & 11.83 & 20.57 & 87.08  & 18.47 & 35.52  & 21.21 & 37.37 & 209.62 & 31.85  \\
      DACC               & 24.34 & 20.13 & 23.07 & 101.75 & 20.40 & 40.87  & 34.35 & 40.59 & 240.48 & 33.88  \\ 
      Power Balance      & \ul{16.28} & \ul{8.93}  & \ul{17.07} & 71.09  & \ul{15.40} & \ul{18.45}  & \ul{11.05} & \ul{27.22} & 163.90 & \ul{21.57}  \\ 
      \midrule
      WN-60f (Ours) & {\bf 11.44} & {\bf 6.32}  & {\bf 15.34} & 71.18 & {\bf 14.44} & {\bf 16.07} & {\bf 9.64}  & {\bf 26.46} & {\bf 162.61} & {\bf 21.29} \\
      Stable Match (\%)   & 99.10 & 99.40 & 99.40 & 99.50 & 99.80 & 98.10 & 99.00 & 98.00 & 93.90  & 98.60 \\ 
      \midrule
      Win (\%)            & {\bf 25.90} & {\bf 29.70} & {\bf 12.20} & 0.00  & {\bf 9.50}  & {\bf 14.60} & {\bf 25.30} & 9.30  & 0.10   & 8.00  \\
      Tie (\%)            & 67.30 & 60.60 & 83.20 & 96.70 & 86.20 & 72.30 & 54.70 & 78.10 & 89.20  & 80.80 \\
      Loss+Unstable (\%)   & 6.80  & 9.70  & 4.60  & {\bf 3.30}  & 4.30  & 13.10 & 20.00 & {\bf 12.60} & {\bf 10.70}  & {\bf 11.20} \\
      \bottomrule
    \end{tabular}
    }
    \caption{Average $\SEq$ ($\downarrow$), success rate of stable matching, and winning/losing rate against the best algorithmic method (underlined). }
    \vspace{1.0em}
    \label{table:SEqCost}
    
    \scalebox{1.00}{
    \small
    \begin{tabular}{lrrrrr|rrrrr} \toprule
      Agents ($N\times M$) & \multicolumn{5}{c}{$20 \times 20$ } & \multicolumn{5}{|c}{$30 \times 30$ } \\ 
      Datasets (Distribution Type) & \multicolumn{1}{c}{UU}    & \multicolumn{1}{c}{DD}    & \multicolumn{1}{c}{GG}    & \multicolumn{1}{c}{UD}     & \multicolumn{1}{c}{Lib}    & \multicolumn{1}{|c}{UU}    & \multicolumn{1}{c}{DD}    & \multicolumn{1}{c}{GG}    & \multicolumn{1}{c}{UD}     & \multicolumn{1}{c}{Lib} \\ \midrule
      GS                 & 89.14 & 146.16 & 108.36  & \ul{\bf 140.53} & 68.62 & 184.05 & 322.05 & 225.49 & \ul{\bf 312.12} & 137.59 	 \\ 
      PolyMin            & 74.19  & 140.99 & 108.04 & 145.28 & 66.94 & 144.48 & 306.28 & 224.13 & 324.54 & 130.79 	\\
      DACC               & 78.49  & 146.71 & 110.06 & 151.34 & 68.75 & 150.71 & 316.18 & 227.52 & 337.43 & 133.59 	\\ 
      Power Balance      & \ul{73.28}  & \ul{140.12} & \ul{106.92} & 140.55 & \ul{65.89} & \ul{138.04} & \ul{302.30} & \ul{\bf 220.26} & {\bf 312.12} & \ul{\bf 126.96} 	\\ 
      \midrule
      WN-60b (Ours) & {\bf 71.29} & {\bf 138.57} & {\bf 106.50} & 140.72 & {\bf 65.63}  & {\bf 137.70}  & {\bf 301.08}  & 221.12  & 313.11  & 127.12 \\
      Stable Match (\%)   & 98.00 & 99.10  & 98.60  & 99.80  & 99.10  & 97.00   & 98.60   & 93.70   & 98.80   & 98.00 \\ \midrule
      Win (\%)            & {\bf 20.60} & {\bf 29.80}  & {\bf 9.40}   & 0.00   & 6.60   & 10.40   & {\bf 25.70}   & 6.20    & 0.10    & 5.60 \\
      Tie (\%)            & 69.50 & 63.70  & 81.90  & 95.00  & 86.60  & 71.60   & 63.20   & 69.50   & 83.90   & 81.50 \\
      Loss+Unstable (\%)   & 9.90  & 6.50   & 8.70   & {\bf 5.00}   & {\bf 6.80}   & {\bf 18.00}   & 11.10   & {\bf 24.30}   & {\bf 16.00}   & {\bf 12.90} \\
      \bottomrule
    \end{tabular}
    }
  \caption{Average $Bal$ ($\downarrow$), success rate of stable matching, and winning/losing rate against the best algorithmic method (underlined).}
  \label{table:BalCost}
     \end{center}
\end{table*}

\subsection{Comparison with Algorithmic Methods}\label{ss:exp2}
{\bf GS} is the better result of applying the GS algorithm \cite{gale1962college} to prioritize each side once. {\bf PolyMin} is an algorithm that minimizes the regret and egalitarian costs (Table \ref{tab:costs}) instead of $\SEq$ and $Bal$ costs, which we can calculate in polynomial time.
{\bf DACC} \cite{dworczak2016deferred} is an approximate algorithm that runs in $O(n^4)$. {\bf PowerBalance} is the state-of-the-art method that runs in $O(n^2)$. 

{\bf WN-60f/b} is WeaveNet with $L=60$ layers. It was trained with samples of $N=30$ and tested on $N=20,~30$. 
Note that we used the asymmetric variant for UD and Lib. See \ref{ss:exp2_supp} for a detailed ablation study for the symmetric and asymmetric variants and so on.

We show the results in Tables \ref{table:SEqCost} and \ref{table:BalCost}.
In the $\SEq$ experiment, the proposed WN-60f achieved the best scores almost in all the cases, and narrowly lost in UD for $N=20$.
In the $Bal$ experiment, WN-60b also achieved the best scores in UU, DD, GG, and Lib for $N=20$ and UU, DD for $N=30$, and narrowly lost in the other four cases.
Note that these scores of WN-60f/b involve unstable matchings.

To evaluate the results while strictly forbidding unstable matchings, we compared $\SEq$ and $Bal$ obtained by WN-60f/b with the best algorithmic method (underlined in the table) sample by sample, and counted win, tie, and loss percentages of WN, where unstable matchings were treated as losses.
Even with this setting, WN-60f beat the best algorithmic method in six among all the ten cases, and WN-60b won four among ten. To our best knowledge, other learning-based methods have never achieved such comparable results.

It is noteworthy that the asymmetric variant achieved a similar performance to the best algorithmic methods even in UD, the most biased setting, and Lib, the simulation of real preference distribution while keeping its parameter efficiency. 

\begin{table}
\centering
{    \small
    \setlength\tabcolsep{4pt} 
    \begin{tabular}{lrr}
    \toprule
        \multicolumn{1}{c}{$100\times100$, UU} & \multicolumn{1}{c}{$\SEq$} & \multicolumn{1}{c}{$Bal$}  \\
        \midrule
        GS  & 1259.39 & 1709.53 \\  
        PolyMin  & 153.35 & 952.85 \\  
        DACC  & 194.65 & 988.02 \\  
        Power Balance & {\bf 49.41} & {\bf 909.73} \\  
        \midrule
        WN-80f/b+Hungarian & 68.36 & 919.75     \\  
        Stable Match (\%) & 89.4 & 80.8 \\
        \bottomrule
    \end{tabular}
}
    \caption{Average $\SEq$ ($\downarrow$) and $Bal$ ($\downarrow$) at $N=100$.}
    \label{tab:N100_score}
\end{table}


\subsection{Demonstration with \texorpdfstring{$N=100$}{N=100}}
We further demonstrate the capability of WeaveNet under relatively large size of problem instances, $N=100$. In this case, we found that the binarization process ${\rm argmax}(\hat{m})$ of WeaveNet does not always successfully yield a one-to-one matching. Specifically, WN-80f and WN-80b found stable matchings for 84.4\% and 73.2\% samples while failed to yield one-to-one matchings for 13.4\% and 19.8\%, respectively (see the Table \ref{tab:N100_bp} in \ref{ss:exp2_supp} for details). To compensate for this problem, we applied the Hungarian algorithm to $\hat{m}$ to force an output of one-to-one matching. The results in Table \ref{tab:N100_score} demonstrate that WeaveNet still achieved a close performance to the state-of-the-art algorithmic method, under a problem size that other learning-based-method studies have never set foot into. We consider this result promising, and expect to further improve the scalability and the binarization reliability of WeaveNet.

%% file: sections/6_conclusion.tex
\section{Conclusion}
This paper proposed a novel neural network architecture, WeaveNet, and evaluated it on fair stable matching, a strongly NP-hard problem. 
It outperformed any other learning-based methods by a large margin and achieved a comparative performance with the state-of-the-art algorithms under a limited size of problem instances.
We hope that our proposed core module, the feature weaving layer, opens a new vista for machine learning and combinatorial optimization applications.

%% file: sections/Z_supplementary.tex
\appendix

\section*{Appendix}

\section{Further Implementation Details}
\subsection{Scaling Function} \label{ss:sf}
$p^A_{ij}$ has a value range of $[1, N]$, which depends on the size of the input problem instance. Moreover, general deep-learning frameworks (e.g., PyTorch and TensorFlow) initialize model weights under the assumption where the maximum value in the input data is around $1.0$.
Hence, to re-scale $p^A_{ij}$ to $s^A_{ij}$, we used the following function $h$, which is
\begin{equation}
\begin{split}
    h(p^A_{ij}) = ((1-\lowest)(N-p^A_{ij}))/N + \lowest \\
    h(p^B_{ji}) = ((1-\lowest)(N-p^B_{ji}))/N + \lowest, \\
\end{split}
\end{equation}
where $\lowest$ should be a constant value in range $(0,1)$ and is set to 0.1 in our experiments.


\subsection{Visualization of Set Encoder and Calculation Cost} \label{ss:set_encoder}
The purpose of the set encoder is to embed the feature of an edge $a_i\rightarrow b_j$ while considering the relative relationship among the other out-edges from $a_i$.
Because the number of out-edges is variable, the set encoder should also accept variable-size inputs.
Besides, since the set of edges have no order, the encoder should output a permutation invariant result.

\begin{figure}[ht!]
    \centering
    \includegraphics[width=1.0\linewidth]{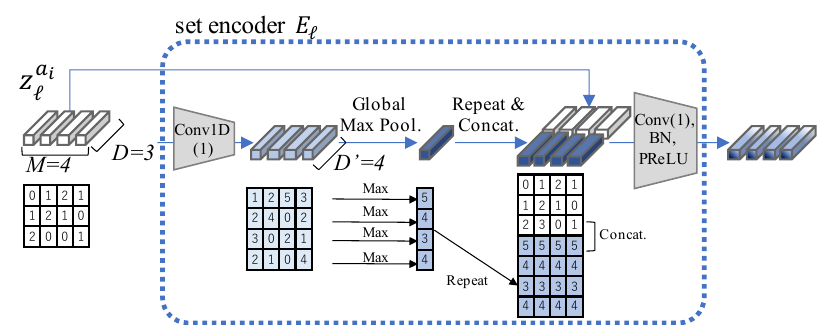}
    \caption{Illustration of the process in set encoder $E_\ell$, where $z_\ell^{a_i}$ (colored in white) is once encoded to $D'$ channel features (colored in pale blue), then max-pooled to obtain statistics in the feature set (colored in blue). The statistics information is concatenated to each input feature and further encoded (color in a gradation).}
    \label{fig:conv_with_global_max_pooling}
\end{figure}

Such network structure is proposed in \cite{deepset} and \cite{pointnet}. Following these studies, we implemented the encoder of WeaveNet as a structure that has two convolutional layers with a set-wise max-pooling layer. The pooling is done for each element in the feature vector and yields a single vector, which is then concatenated to each input feature. The illustration is in Fig.~\ref{fig:conv_with_global_max_pooling}.
In the experimental part, we refer to the number of output channels from the convolutional layers as $D'$ and $D$, where $D'$ is the intermediate feature fed to the max-pooling layer and $D$ is the output of this encoder. Following \cite{pointnet}, $D'$ is typically set larger than $D$.
Note that the structure is followed by the batch normalization and activation, for which we adopted the PReLU function. 

\subsubsection{Computational Complexity}
With $D_{\max} = \max\{D',D\}$, the theoretical calculation cost of a single set encoder on a single core processing unit is $O(N^2D_{\max}^2)$, where a single linear operation in the convolution requires $D^2$ and we repeat it for $2N\times M$ elements ($N\geq M$).
The other operations in WeaveNet, such as cross-concatenation, clearly have less cost. Hence, the entire WeaveNet has its calculation cost of $O(LN^2D_{\max}^2)$.

We can reduce this into $O(L\log(ND_{\max}))$ with an ideal parallel processing unit.
We have $2N\times M$ elements for the convolution and can linearly operate them independently. Each linear operation can be done in parallel except for weighted sum aggregation, which takes $O(\log(D_{\max}))$. Besides, we have a max-pooling operation, which consists of $D$ parallel max operations. It can also be done in parallel and each max operation for $N$ elements takes $O(\log N)$. Therefore, we can execute the calculation of a single set encoder in $O(\log(N)+\log(D_{\max}))$.

\subsection{Information Propagation with WeaveNet}\label{app:scope}
\begin{figure}[t!]
    \centering
    \includegraphics[width=1.0\linewidth]{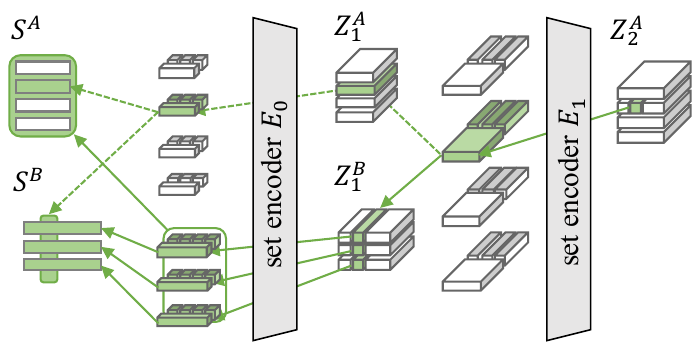}
    \caption{Backward path of WeaveNet (two-stream architecture) from $ij$-th feature in $Z^A_2$. The upper stream terminates at $i$-th row of $S^A$ and $i$-th column of $S^B$, which are the weights on outgoing and incoming edges of the agent $a_i$, respectively. 
    The lower stream refers to all the elements in $S^A$ and $S^B$.}
    \label{fig:backward_path}
\end{figure}
\begin{figure}[t!]
    \centering
    \includegraphics[width=1.0\linewidth]{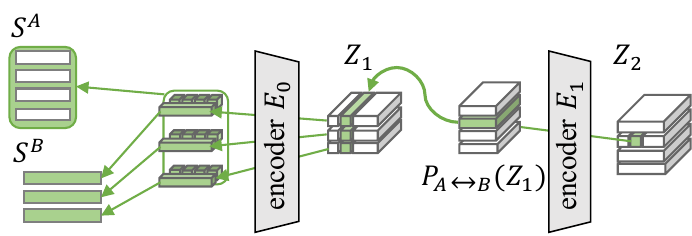}
    \caption{Backward path of DBM (single-stream architecture) from $ij$-th feature in $Z_2$, where the stream refers to all the elements in $S^A$ and $S^B$. This structure is common with SSWN.}
    \label{fig:backward_path_dbm}
\end{figure}

\begin{figure*}[!t]
    \centering
    \begin{minipage}{0.5\textwidth}
    \includegraphics[width=0.9\linewidth]{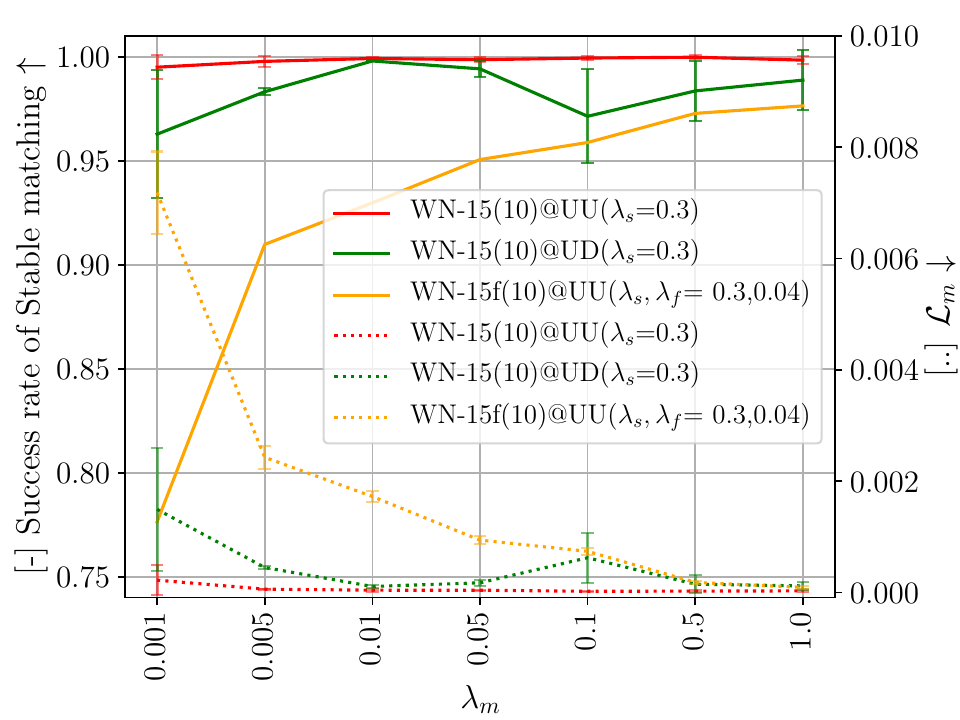}
    \vspace{-0.5em}
    \caption{Sensitivity against $\lambda_m$}
    \label{fig:suppli_c}
    \end{minipage}%
    \begin{minipage}{0.5\textwidth}
    \includegraphics[width=0.9\linewidth]{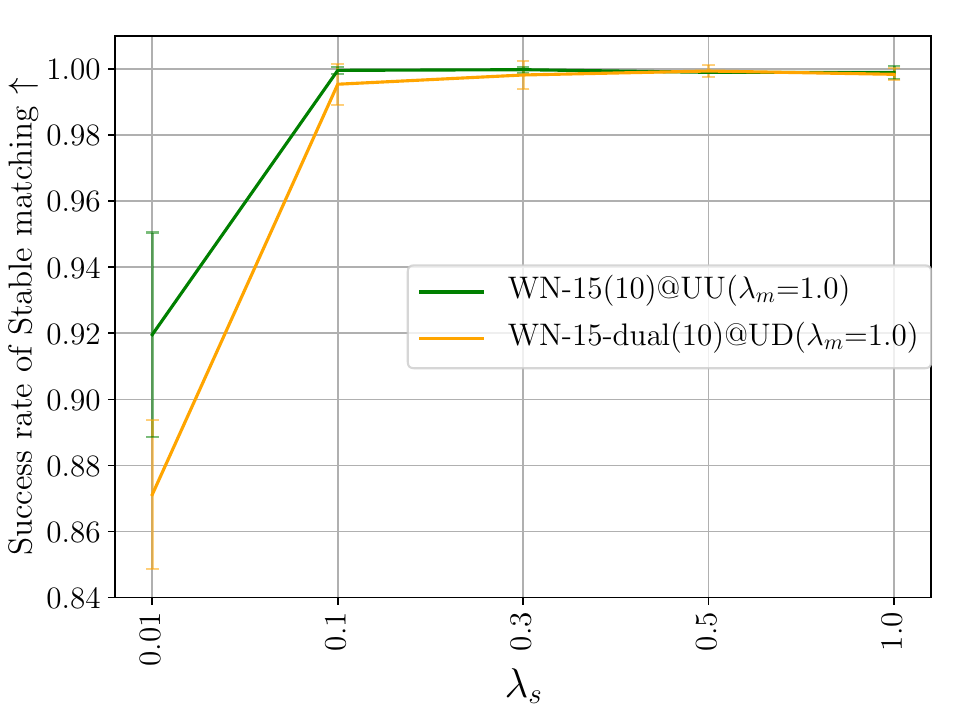}
    \vspace{-0.5em}
    \caption{Sensitivity against $\lambda_s$}
    \label{fig:suppli_s}
    \end{minipage}\\
    \begin{minipage}{0.5\textwidth}
    \includegraphics[width=0.9\linewidth]{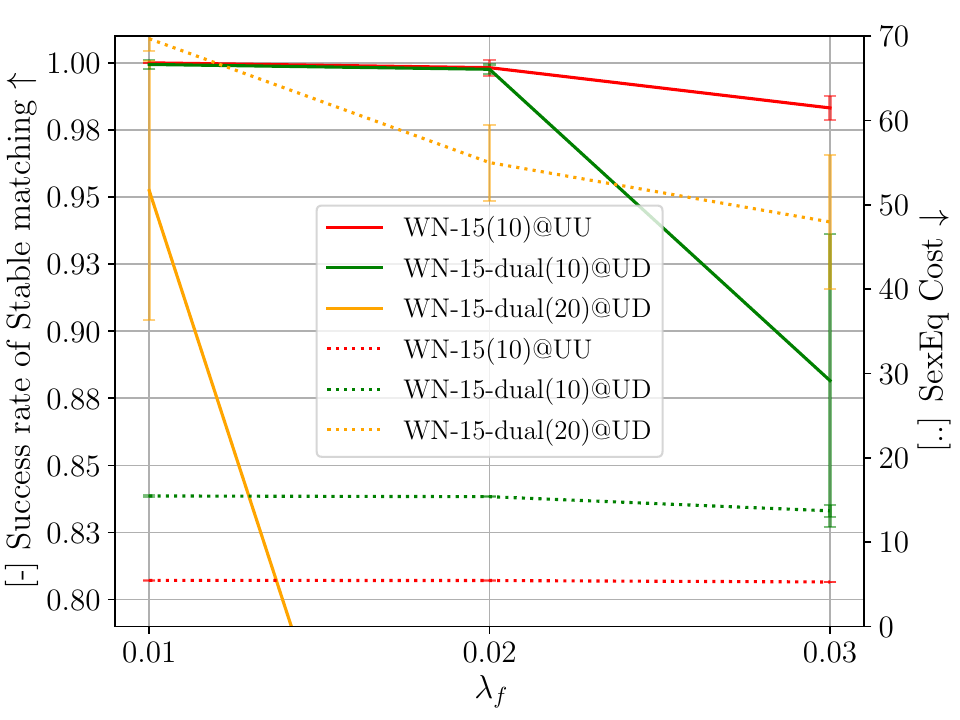}
    \vspace{-0.5em}
    \caption{Sensitivity against $\lambda_f$}
    \label{fig:suppli_f}
    \end{minipage}%
    \begin{minipage}{0.5\textwidth}
    \includegraphics[width=0.9\linewidth]{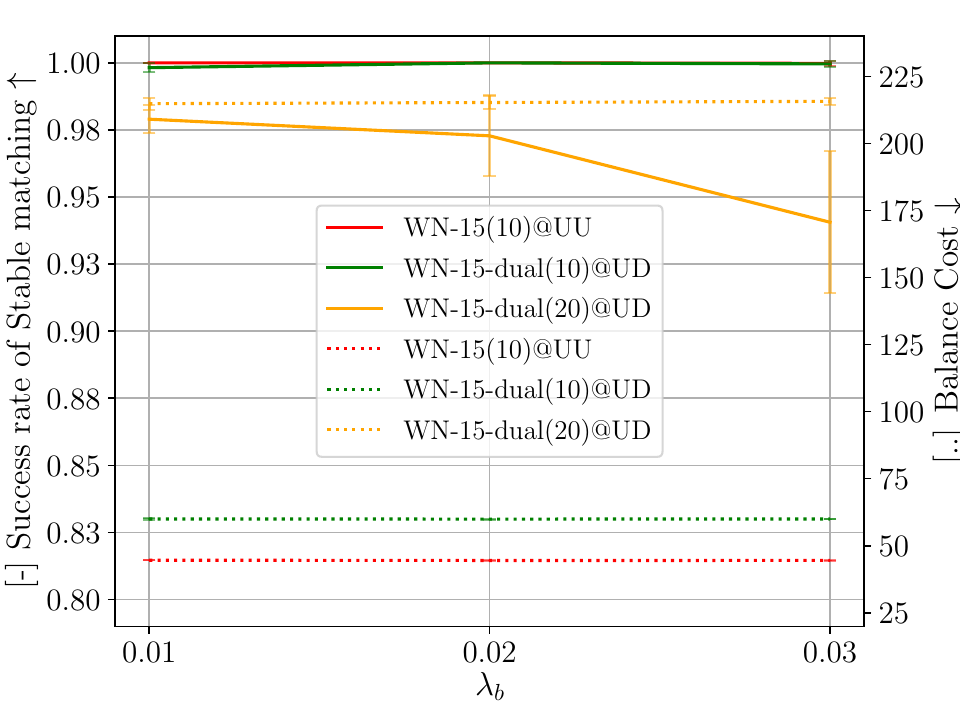}
    \vspace{-0.5em}
    \caption{Sensitivity against $\lambda_b$}
    \label{fig:suppli_b}
    \end{minipage}
\end{figure*}

\checkedby{Jiaxin}{Hashi}
By stacking two or more FW layers, every latent feature of $ij$-th element in $Z^*_\ell$ ($\ell\geq 2$) has a receptive field that covers all over the preference lists.
Fig.~\ref{fig:backward_path} illustrates the receptive field, where green elements are upstream components of the $ij$-th element, and $E_0$, $E_1$ are encoders that involve all the input into the calculation of each feature in the output sequence. 
Let us backtrack the path of back-propagation from the $ij$-th element in $Z^A_2$.
It derives from the $i$-th row of $Z^A_1$ and the $i$-th column of $Z^B_1$.
Focusing on the $i$-th column (e.g., the green column) of $Z^B_1$, each element (in the $k$-th row) derives from all the elements in the $k$-th row of $S^B$, plus all the elements in the $k$-th column of $S^A$.
In this way, all the elements in $S^A$ and $S^B$ contribute to a column of $Z^B_1$, then an element of $Z^A_2$.
Symmetrically, all the elements in $S^A$ and $S^B$ contribute to a column of $Z^A_1$, then an element of $Z^B_2$.
Hence, we can see that stacking two FW layers can cover the entire bipartite graph in its reference.

We also visualize the back-propagation path of DBM in Fig.~\ref{fig:backward_path_dbm}.
DBM also covers the entire bipartite graph within two layers, but each encoder is used only for one direction (either weftwise or warpwise). Here, the path in Fig.~\ref{fig:backward_path_dbm} is identical with the lower path in Fig.~\ref{fig:backward_path}. From this perspective, WeaveNet innately contains the path of DBM.

\subsection{Loss Weights}\label{app:hyperparameters}

Through the experiments, we set the loss weights $\lambda_m=1.0$, $\lambda_s=0.7$, and $\lambda_f=\lambda_b=0.01$ to train any learning-based methods.
We adjusted these parameters in the following process.

As a preliminary model, we prepared WN-15 with a set encoder with $D=64$ and $D'=256$. In this investigation, we used a balanced dataset UU and the most biased dataset UD.

Because we experimentally found the tendency that the model hardly outputs a stably matched solution without minimizing $\loss_m$, we first tried to fix $\loss_m$.
Fig.~\ref{fig:suppli_c} shows the success rate of stable matching with different $\loss_m$ in the range $\{0.001,~0.005,~0.010,~0.050,~0.100,~0.500,~1.000\}$, where WN-15($n$) is the model trained and validated with the samples of $N=n$. 
From this result, we decided to set $\lambda_m=1.0$ and use it as the maximum weight among the loss weights.

Next, fixing $\lambda_m=1.0$, we observed the trend in success rate of stable matching against $\lambda_s$. Fig.~\ref{fig:suppli_s} shows our investigation of $\loss_s$ in the range $\{0.01,0.10,0.30,0.50,1.00\}$. 
Here, WN-dual is a WeaveNet variant to deal with the inconsistent distributions of UD (see \ref{ss:exp2_supp} for details of WN-dual).
As a result, we found that $\lambda_s$ should simply be large enough (ca. $\lambda_s\geq 0.30$).

To investigate sensitivity of the model against $\lambda_f$ and $\lambda_b$, in this experiment, we set $\lambda_s=0.3$ as the minimum satisfiable value in Fig.~\ref{fig:suppli_s}\footnote{We used $\lambda_s=0.7$ in any other experiments to stabilize the results, as stated in Section \ref{s:exp}.}, and obtained Figures \ref{fig:suppli_f} and \ref{fig:suppli_b}. From these figures, we found that too-strong weights for fairness may disrupt stability, which matches a theoretical expectation. Hence, we decided to use $\lambda_f=\lambda_b=0.01$, which achieved the highest success rate of stable matching in the search range of $\{0.01,0.02,0.03\}$.

\subsection{Network Architecture}\label{ss:archi}
We can customize the shape of WeaveNet architecture finely by using set encoders with different shapes in each FW layer; however, we avoided such a complex architecture design to simplify the analysis and decided to use common-shape set encoders for all $L$ FW layers.
With this setting, we have only three hyper-parameters to decide the architecture: $L$, $D$, and $D'$. The shortcut path is regularly connected to the input of $\ell$-th layer with an even number of $\ell$.

\begin{table}[ht]
    \centering
{    \small
     \setlength\tabcolsep{2pt} 
    \begin{tabular}{lrrrrr}
    \toprule
        \multicolumn{1}{c}{Name} & \multicolumn{1}{c}{$L$} & \multicolumn{1}{c}{$D$} & \multicolumn{1}{c}{$D'$} & \multicolumn{1}{c}{\# of params.} & \multicolumn{1}{c}{Stably Matched (\%)} \\
        \midrule
        Deep  & 30 & 22 & 44 & 117k & 95.7\% \\  
        Wide1 & 15 & 32 & 64 & 119k & 76.0\% \\ 
        Wide2 & 15 & 24 & 98 & 120k & 73.1\% \\ 
        \bottomrule
    \end{tabular}
}
    \caption{Comparison in parameter efficiency among different shape architectures. Deep achieved the best success rate in stable matching despite its smallest architecture.}
    \label{tab:supp_architecture}
\end{table} 

Table \ref{tab:supp_architecture} shows the success rate of stable matching after 100,000 iterations of training. Here, we drew samples from the UU distribution with $N=30$ for both training and validation.
The result shows that the deepest model performs the best despite its smallest number of parameters. Hence, we decided to use the set encoder that has $D\leq 32$ and $D'=2D$.

\begin{figure}[ht]
    \centering
    \includegraphics[width=1.0\linewidth]{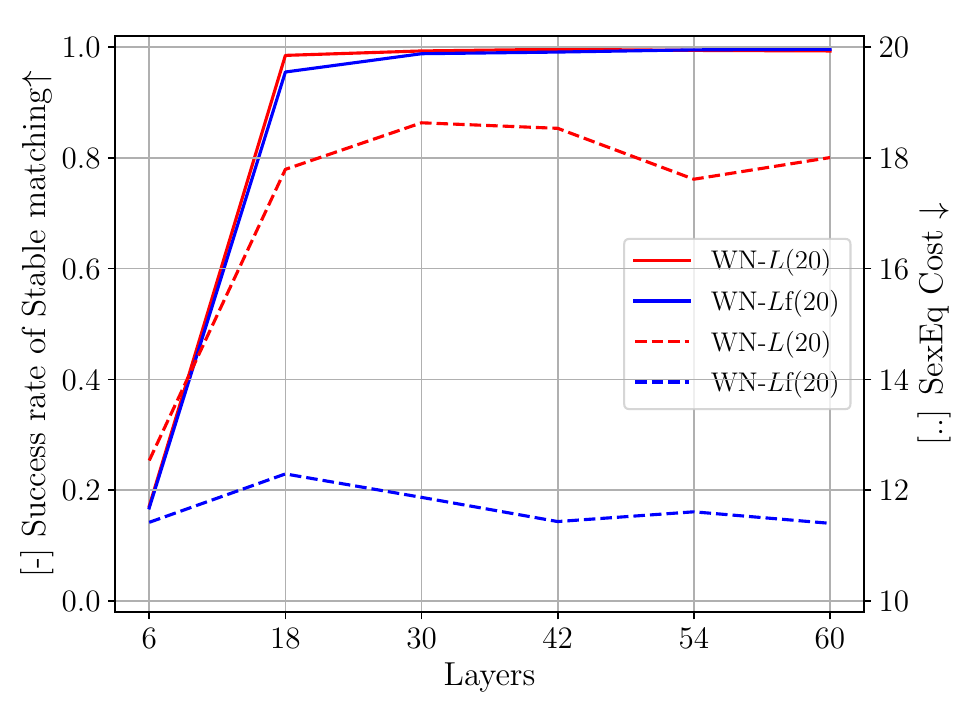}
    \caption{Success rate of stable matching (solid, $\uparrow$) and $\SEq$ (dashed, $\downarrow$) according to $L$.}
    \label{fig:impact_of_depth}
\end{figure}

We further investigated the impact of network depth on the problem.
To observe how the strongly NP-hard target increases the difficulty in optimization, we plotted both results by WN-$L$ and WN-$L$f with $L\in \{6,18,30,42,54,60\}$. 
Fig.~\ref{fig:impact_of_depth} shows the trend of success rate against different $L$. Here, the models were trained and validated with the UU dataset at $N=20$.
We can see from the result that $L=6$ is not enough to stably match samples of $N=20$, but $L=18$ is enough if only for that purpose.
Besides, we observed that a deeper stack of layers tends to improve $\SEq$ slightly.

\begin{table}[!ht]
    \centering
     \setlength\tabcolsep{4pt} 
    {\small
    \begin{tabular}{lrrrrr}
    \toprule
        \multicolumn{1}{c}{Model} & \multicolumn{1}{c}{Encoder} &  \multicolumn{1}{c}{$D$} & \multicolumn{1}{c}{$D'$} & Res. &\# of params. \\
        \midrule
        MLP-3  & dense layer & 100 & - & w/o & 28k  \\ 
        GIN-2 & graph conv. & 44 & - & w/o & 29k \\ %
        DBM-6 & max pooling & 48 & - & w/o & 29k  \\ 
        DBM\_A-6 & self-attention & 48 & 32 & w/ & 28k\\ %
        SSWN-6 & set encoder  & 48 & 48 & w/o & 25k \\ 
        SSWN-60 & set encoder & 64 & 64 & w/ & 740k \\ 
        WN-6 & set encoder & 24 & 48 & w/o & 25k \\ 
        WN\_A-6 & self-attention & 32 & 28 & w/ & 29k \\ %
        WN-18 & set encoder & 32 & 64 & w/ & 143k \\ 
        WN-60 & set encoder & 32 & 64 & w/ & 493k \\ %
        WN-80 & set encoder & 32 & 64 & w/ & 659k \\
        \bottomrule
    \end{tabular}
    }
    \caption{Hyper-parameters for each baseline, where $D'$ represents the length of key and query features for self-attention, and the max-pooling input for the set encoder. Note that the $D$-dimentional feature of WN is cross-concatenated and processed as a $2D$-dimensional features in the encoders. }\label{tab:hypara_baselines}
\end{table} 

\subsection{Hyper-parameter Settings for the Other Baselines}\label{ss:supp_hypara_baseline}

Here, we denote the hyper-parameters for learning-based baseline methods, including those used in \ref{ss:exp1_supp} and \ref{ss:exp2_supp}. 
The detail is summarized in Table \ref{tab:hypara_baselines}. Note that MLP could not converge when more than three hidden layers are stacked. Hence, we set the number of layers to be three, as used in \cite{harvard2019}. For GIN \cite{xu2018how}, we put two layers because the original paper reported it performs best. Besides that, some papers provide theoretical analyses that graph convolutional networks (GCNs) suffer from over-smoothing \cite{li2018deeper,Oono2020Graph}. Namely, it loses the discriminative capacity for nodes on a dense graph when the layers are deeply stacked. This is exactly our case because a complete bipartite graph is always dense; its connection density is $N/(2N-1)(>1/2)$ (there are $2N^2$ edges among $2N(2N-1)$ possible edges). Hence, a deeper stack does not contribute to solving the problem.

Note that we used the same loss weights decided in \ref{app:hyperparameters} for all the methods since the loss weights derive from the task property rather than the architecture.
For simplicity in comparison, we also used the three hyper-parameters, $L$, $D$, and $D'$, to identify the architecture following to \ref{ss:archi}. We experimentally decided whether to use the residual structure or not for the models which have more than four layers. Hence, the shown results are always better ones.

\section{Additional Experiments}\label{ss:supp_additional_exp}
\subsection{Additional Learning-based Baselines and Ablations.}\label{ss:exp1_supp}

\begin{figure}[th]
    \centering
    \includegraphics[width=1.0\linewidth]{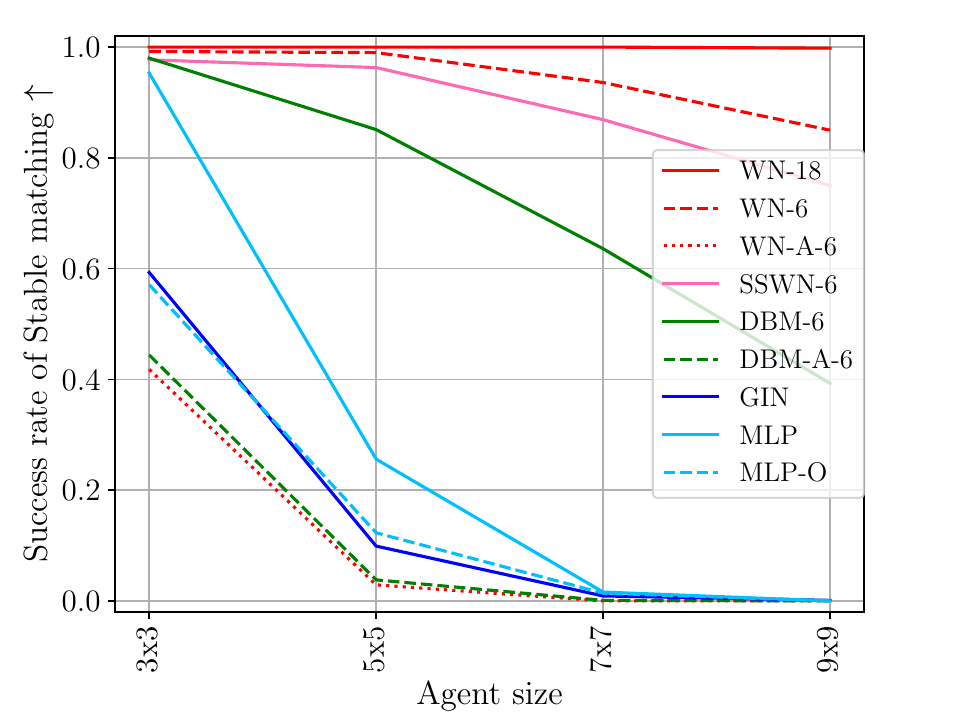}
    \caption{Change in success rate of stable matching ($\uparrow$) according to $N$, as a complement of Fig.~\ref{fig:Ex1_a}.}
    \label{fig:exp1_supp}
\end{figure}

We compared WeaveNet to more comprehensive baselines: a variation of MLP, DBM, WeaveNet, and a GCN model.
Fig.~\ref{fig:exp1_supp} shows the results.

{\bf MLP-O} is a variant of MLP that faithfully follows the loss functions proposed in \cite{harvard2019}.
The difference is in the matrix constraint loss function $\loss_m$.
Ours defined in Eq.~\eqref{eq:loss_c_p2} is cosine-distance based, while the one proposed in the original paper is Euclidean-distance based, which is
\begin{equation}
\begin{split}
\loss_m(\hat{m}^A,\hat{m}^B) & = \sum_{\{i,j\}\in A\times B} |\hat{m}^A_{ij} - \hat{m}^B_{ji} |.
\end{split}
\label{eq:loss_c_orig}
\end{equation}

The above loss function restricts $i$-th column and $i$-th row of $\hat{m}$ to be the same point in the $N\times M$ space. In other words, the column and row must match in a scale variant manner.
We considered that this is unnecessarily strict for maintaining $\hat{m}$ to be symmetric. Hence we adopted the cosine distance for $\loss_m$. 
In Fig.~\ref{fig:exp1_supp}, we observed that MLP has a clear performance gain from MLP-O by our cosine-distance-based $\loss_m$.

{\bf GIN} is the state-of-the-art GCN architecture \cite{xu2018how}.
Some readers may consider GCNs as the solution to this problem. 
We put two GIN layers followed by one Linear layer that outputs $\hat{m}$ as a flat vector as MLP. From the result, we confirmed that the GCN is not suitable to solve this problem.

DBM proposed in \cite{gibbons2019deep} has two variations; one uses max-pooling with one convolutional layer\footnote{The max-pooled feature is copied and concatenated to the input feature as the set encoder, but it does not have convolution before the max-pooling operation.} as the layer-wise encoder (DBM) and self-attention (DBM\_A). 
We also integrated self-attention into the WeaveNet architecture ({\bf WN\_A-6}) to complete the comparison of the encoder choice. 
As a result, we have confirmed that self-attention does not work well for stable matching problem.

\begin{table*}[th!]
\setul{1pt}{.4pt}
\centering
{\small

\begin{tabular}{lrrrrr|rrrrr}
\toprule
Agents ($N\times M$) & \multicolumn{5}{c|}{20$\times$20}  & \multicolumn{5}{c}{30$\times$30}  \\ \midrule
      Datasets (Distribution Type)     & \multicolumn{1}{c}{UU}    & \multicolumn{1}{c}{DD}    & \multicolumn{1}{c}{GG}    & \multicolumn{1}{c}{UD}     & \multicolumn{1}{c}{Lib}    & \multicolumn{1}{|c}{UU}    & \multicolumn{1}{c}{DD}    & \multicolumn{1}{c}{GG}    & \multicolumn{1}{c}{UD}     & \multicolumn{1}{c}{Lib} \\ \midrule
WN-60f-sym             & \ul{11.44}  & \ul{6.32} &  \ul{15.34} & 81.21 & 14.39 & \ul{16.07} & \ul{9.64}  & \ul{26.46} & 194.24  & 21.16 \\
Stably Matched (\%)  & \ul{99.10} & \ul{99.40}  & \ul{99.40} & 96.00  & 99.50 & \ul{98.10}  & \ul{99.00}  & \ul{98.00} & 81.50  & 98.80 \\ \midrule
WN-60f-asym          & -& -& -& \ul{71.18} & \ul{14.44} & -& -& -& \ul{162.61} & \ul{21.29} \\
Stably Matched (\%)  & -& -& -& \ul{99.50} & \ul{99.80} & -& -& -& \ul{93.90} & \ul{98.60}\\ \midrule
WN-60f-dual          & -& -& -& 71.25 & 14.58 & - & - & - & 163.71 & 22.12\\
Stably Matched (\%)  & -& -& -& 98.50 & 99.60 & - & - & - & 94.80  & 97.50 \\ \midrule
WN-60f(20)          & 11.49 & 6.20 & 15.30 & 71.10 & 14.40 & 16.66 & 9.55 & 26.57 & 162.91 &  21.82 \\
Stably Matched (\%) & 98.90 & 99.50 & 99.40 & 99.60 & 99.30 & 94.60 & 97.30 & 95.70 & 91.30 & 97.70 \\ \midrule
SSWN-60f          &  12.82 & 7.41 & 16.62 & 70.94 & 15.20 & 19.13 & 12.76 & 28.33 & 160.83 & 22.45 \\
Stably Matched (\%)  & 95.80 & 99.20 & 96.50 & 92.40 & 97.90 & 90.00 & 97.50 & 89.00 & 69.20 & 92.30  \\ \midrule \midrule
WN-60f+Hung &   11.51 &	6.35 & 15.37 &	71.18 &	14.45 &	16.14 &	9.69 & 26.58 & 	162.73 & 21.32  \\ 
Stably Matched (\%)  &  99.80 & 99.90 & 100.00 & 99.60 & 100.00 & 99.70 & 99.90 & 99.90 & 95.20 & 99.20 \\ \bottomrule
\end{tabular}
}
\caption{Ablation study for the network architecture difference with $\SEq$ ($\downarrow$) and the success rate of stable matching ($\uparrow$). The scores shown in Table \ref{table:SEqCost} are underlined.}
\end{table*}

\begin{table*}[th!]
\setul{1pt}{.4pt}
\centering
  {\small

\begin{tabular}{lrrrrr|rrrrr}
\toprule
Agents ($N\times M$) & \multicolumn{5}{c|}{20$\times$20}  & \multicolumn{5}{c}{30$\times$30}  \\ \midrule
Datasets (Distribution Type) & \multicolumn{1}{c}{UU}    & \multicolumn{1}{c}{DD}    & \multicolumn{1}{c}{GG}    & \multicolumn{1}{c}{UD}     & \multicolumn{1}{c}{Lib}    & \multicolumn{1}{|c}{UU}    & \multicolumn{1}{c}{DD}    & \multicolumn{1}{c}{GG}    & \multicolumn{1}{c}{UD}     & \multicolumn{1}{c}{Lib} \\ \midrule
WN-60b-sym              & \ul{71.29} & \ul{138.57} & \ul{106.50} & 141.96 & 65.59 & \ul{137.70} & \ul{301.08} & \ul{221.12} & 315.51 & 127.05 \\
Stably Matched (\%)     & \ul{98.00} & \ul{99.10}  & \ul{98.60}  & 97.30  & 98.90 & \ul{97.90}  & \ul{98.60}  & \ul{93.70}  & 80.30  & 98.10 \\ \midrule
WN-60b-asym          & -& -& -& \ul{140.72} & \ul{65.63} & -& -& -& \ul{313.11} & \ul{127.12} \\
Stably Matched (\%)  & -& -& -& \ul{99.80} & \ul{99.10} & -& -& -& \ul{98.80} & \ul{98.00}\\ \midrule
WN-60b-dual          & -& -& -& 141.20 & 65.70 & - & - & - & 314.79 & 127.25 \\
Stably Matched (\%)  & -& -& -& 99.60 & 99.30 & - & - & - & 98.90 & 97.90 \\ \midrule
WN-60b(20)          & 71.05 & 138.53 & 106.13 & 140.81 & 65.53 & 137.06 & 301.21 & 220.23 & 313.68 & 127.62 \\
Stably Matched (\%)  & 98.50 & 98.80 & 99.50 & 99.70  & 98.80 & 96.10 & 96.70 & 95.00 & 88.90 & 93.80 \\ \midrule
SSWN-60b             & 76.54 & 139.29 & 107.58 & 141.37 & 66.99 & 150.88 & 302.90 & 223.62 & 314.91 & 129.52 \\
Stably Matched (\%)  & 92.60 & 98.50 & 98.40 &  99.60 & 98.60 & 79.80 & 96.70 & 93.90 & 94.0 & 92.80 \\ \midrule \midrule
WN-60b+Hung & 71.45 & 138.60 & 106.55 & 140.73 & 65.68 & 137.92 & 301.14 & 221.42 & 313.13 & 127.22 \\ 
Stably Matched (\%)  & 99.90 & 99.90 & 99.60 & 99.90 & 100.00 & 99.00 & 99.90 & 96.90 & 99.30 & 99.20 \\ \bottomrule
\end{tabular}
}
\caption{Ablation study for the network architecture difference with $Bal$ ($\downarrow$) and the success rate of stable matching ($\uparrow$). The scores shown in Table \ref{table:BalCost} are underlined.}
\end{table*}

\subsection{Additional Ablation in \texorpdfstring{$N=20,~30$}{N=20, 30}} 
\label{ss:exp2_supp}

We provide a detailed ablation study to finely analyze the symmetric and asymmetric variants, the effect of two-stream architecture, the size of training samples, and the binarization operation.

First, we refer to the symmetric and asymmetric variants of WN-60f/b as {\bf WN-60f/b-sym} and {\bf WN-60f/b-asym}, respectively. Note that both models have the two-stream architecture, but {\bf sym} shared batch normalization layers and {\bf asym} does not. In addition, {\bf asym} has an additional channel in inputs for side-identifiable code. 
In addition to them, we further prepared {\bf WN-60f/b-dual}, a model that separately holds the entire set encoder for $Z^A_\ell$ and $Z^B_\ell$ at each layer to deal with asymmetric inputs. 
Since UD is a highly asymmetric dataset, the symmetric variant (WN-60f/b-sym) does not work appropriately while the asymmetric variants ({\it asym} and {\it dual}) could deal with it.
In the dataset Lib, where we assumed an unknown bias strength, {\it asym} worked as well as {\it sym} and {\it dual}. This result showed that we can safely apply the asymmetric variation for any situation.

Comparing {\it asym} with {\it dual}, {\it asym} marked slightly smaller fairness costs than {\it dual} while keeping comparative success rate in stable matching.
This result experimentally confirmed the parameter efficiency of {\it asym} against {\it dual} since {\it dual} has roughly twice more parameters than {\it asym}.

Second, to investigate the effect of choosing different training sample size, we prepared {\bf WN-60f/b(20)}, a model trained by samples of $N=20$, and applied it to test datasets of $N=20,~30$. The asymmetric variant was applied for UD and Lib.
From the results, we confirmed that the models trained with $N=30$ has no degradation from WN-60f/b(20) even when tested at $N=20$. In contrast, WN-60f/b(20) did not perform well when applied to the samples of $N=30$.
This trend implies that we should train the model with a larger $N$ to cover a wide range of input-size variations.

Third, {\bf SSWN-60}, a single-stream WeaveNet with set encoders, was prepared to measure the contribution of the two-stream architecture. 
SSWN-60f/b achieved consistently worse result than WN-60f/b (underlined), and collapsed in UD ($N=30$) in Table~\ref{table:SEqCost} and UU ($N=30$) in Table~\ref{table:BalCost}.
These results experimentally proved the importance of the two-stream structure.

Finally, to avoid an increase of theoretical calculation cost, WeaveNet applied ${\rm argmax}$ operation to binarize $\hat{m}$ rather than the Hungarian algorithm, whose calculation cost is $O(N^3)$. 
To demonstrate how well the network maintains ${\rm argmax}(\hat{m})$ to be a matching, we compared the result with those obtained with the Hungarian algorithm ({\bf WN-60f+Hung}). Again, the asymmetric variant was applied for UD and Lib.
From the result, we confirmed that WN-60f/b+Hung achieved higher success rates in stable matching while worse $\SEq$ and $Bal$ costs. 
This behavior implies that the output matched only by the Hungarian algorithm tends to have higher fairness costs.

To obtain further insight, we prepared Table~\ref{tab:N100_bp}, which summarizes the difference w/ and w/o the Hungarian algorithm at $N=100$. In this setting, we have more samples with which WN-80f/b fails to even obtain one-to-one matching by the ${\rm argmax}$ binarization.
Binarization by the Hungarian algorithm forces such failure estimation $\hat{m}$ to be a matching; however we obtained only 5.0\% additional stable matchings from the 13.4\% failure cases with WN-80f, and 7.5\% from 19.8\% failure cases with WN-80b. A similar number of cases resulted in matchings with more than three blocking pairs. 

From these results, we concluded that the Hungarian algorithm does not essentially improve the quality of fair stable matching, although it ensures $\hat{m}$ to be a one-to-one matching.

\begin{table}
\centering
{    \small
     \setlength\tabcolsep{4pt} 
    \begin{tabular}{rrr|rr}
    \toprule
        \multicolumn{1}{c}{\#Block. Pairs} & \multicolumn{1}{c}{WN-80f} & \multicolumn{1}{c|}{+Hung.} & \multicolumn{1}{c}{WN-80b} & \multicolumn{1}{c}{+Hung.} \\
        \midrule
        0  & 84.4\% & 89.4\%     & 73.2\% & 80.8\% \\  
        1 & 2.2\% & 4.6\%        & 6.7\% & 10.9\% \\ 
        2 & 0.0\% & 0.4\%        & 0.3\% & 1.2\% \\ 
        $\geq 3$ & 0.0\% & 5.6\% & 0.0\% & 7.1\% \\ 
        Fail & 13.4\% & -        & 19.8\% & - \\ 
        \bottomrule
    \end{tabular}
}
    \caption{Number of blocking pairs in the estimated matching with UU $N=100$. Fail counts outputs that are not a one-to-one matching. Outputs with no blocking pairs are stable matchings.}
    \label{tab:N100_bp}
\end{table}

\section{Data for Reproduction}

\subsection{Calculation Time and Computing Infrastructure}
We trained and tested the learning-based models used in the experiments on a single GPU (Tesla V100, memory size 16GB) mounted on NVIDIA DGX-1. We developed the environment on Ubuntu18.04. 
\begin{table}
    \centering
  {\small
    \begin{tabular}{lrr}
    \toprule
        \multicolumn{1}{c}{Elapsed Time} & \multicolumn{1}{c}{for training}  & \multicolumn{1}{c}{for test} \\
        & 200,000 iters. & 1,000 samples \\
        \midrule
        WN-6~ ($N=10$)  & 10.39 hours & 5.82 s \\ 
        WN-18 ($N=10$) & 13.10 hours & 14.83 s \\ 
        WN-60 ($N=20$) & 30.10 hours & 75.15 s \\
        WN-60 ($N=30$) & 43.58 hours & 74.10 s \\ 
        WN-80 ($N=100$) & 111.10 hours & 103.66 s \\ 
    \end{tabular}
    }
    \caption{Elapsed time for training and testing the models.}
    \label{tab:elapsed_time}
\end{table} 

The above training required less than 16GB of memory space in our setting as long as the batch size is no larger than $8$.
The training and inference time are summarized in Table \ref{tab:elapsed_time}.

\subsection{Dataset}
We implemented the data generator for UU, DD, GG, UD, and Lib following the explanation in \cite{tziavelis2019equitable} since they are not provided by the authors. 
In the process, we also extracted the distribution of the LibimSeTi dataset \cite{brozovsky07recommender}, where the original data of the LibimSeTi dataset is accessible in \url{http://www.occamslab.com/petricek/data/}. 
We filtered out data that do not have bidirectional preference rank, as stated in \cite{tziavelis2019equitable}.
We yielded the validation and test datasets with a fixed random seed to make them reproducible. 

All the code for data generation is included in the submission (with the random seeds and fairness costs of each sample obtained by the traditional algorithmic baselines).  They will become publicly available at the timing of publication of this paper.